\pdfoutput=1

\documentclass[11pt]{article}

\usepackage[preprint]{acl}

\usepackage{times}
\usepackage{latexsym}
\usepackage{hyperref}
\usepackage{booktabs}
\usepackage{amsfonts}
\usepackage{xcolor,colortbl}
\definecolor{amethyst}{rgb}{0.6, 0.4, 0.8}
\definecolor{apricot}{rgb}{0.98, 0.81, 0.69}
\definecolor{atomictangerine}{rgb}{1.0, 0.6, 0.4}
\definecolor{orange_assign}{rgb}{0.9686274509803922, 0.8509803921568627, 0.7686274509803922}
\usepackage{array}
\newcolumntype{P}[1]{>{\centering\arraybackslash}p{#1}}
\usepackage{longtable}
\usepackage{makecell}
\usepackage{multirow}
\usepackage{array}
\usepackage[T1]{fontenc}
\usepackage{enumitem}

\usepackage[T1]{fontenc}

\usepackage[utf8]{inputenc}
\usepackage[inkscapelatex=false]{svg}
\usepackage{tablefootnote}

\usepackage{microtype}

\usepackage{inconsolata}
\usepackage{microtype}
\usepackage{amsmath}
\usepackage{listings}
\usepackage{tabularx}
\usepackage{adjustbox}

\usepackage{graphicx}
\usepackage[nolist]{acronym}
\usepackage{xcolor}

\title{\includegraphics[height=1cm]{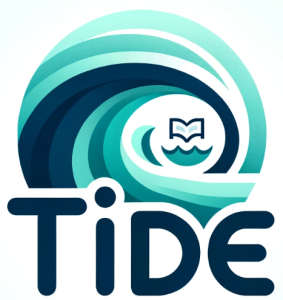} Industry-Aligned Granular Topic Modeling}

\author{
 \textbf{Sae Young Moon\textsuperscript{1}},
 \textbf{Myeongjun Erik Jang\textsuperscript{1}},
 \textbf{Haoyan Luo\textsuperscript{1}},
 \textbf{Chunyang Xiao\textsuperscript{1}},
\\
 \textbf{Antonios Georgiadis\textsuperscript{1}},
 \textbf{Fran Silavong\textsuperscript{1}}
\\
\\
 \textsuperscript{1} J.P. Morgan Chase
\\
 \small{
   \texttt{\{saeyoung.moon, myeongjun.jang, haoyan.luo, chunyang.xiao,}
   }\\ 
 \small{
   \texttt{antonios.georgiadis, fran.silavong\} \href{@jpmchase.com}{@jpmchase.com}}
 }
}

\begin{document}
\maketitle

\begin{abstract}
Topic modeling has extensive applications in text mining and data analysis across various industrial sectors. Although the concept of granularity holds significant value for business applications by providing deeper insights, the capability of topic modeling methods to produce granular topics has not been thoroughly explored. In this context, this paper introduces a framework called TIDE, which primarily provides a novel granular topic modeling method based on large language models (LLMs) as a core feature, along with other useful functionalities for business applications, such as summarizing long documents, topic parenting, and distillation. Through extensive experiments on a variety of public and real-world business datasets, we demonstrate that TIDE's topic modeling approach outperforms modern topic modeling methods, and our auxiliary components provide valuable support for dealing with industrial business scenarios. The TIDE framework is currently undergoing the process of being open sourced.
\end{abstract}

\section{Introduction}
Topic modeling is a \ac{NLP} technique designed to discover meaningful topics within a corpus, which has a broad range of practical usage for text mining and data analysis in research and industry~\cite{ranaei2017topic, asmussen2019smart, xiong2019analyzing}. From a business perspective, granularity — referring to the breakdown of a concept into smaller, more detailed parts~\cite{mulkar-mehta-etal-2011-granularity} — is an important consideration. Higher levels of granularity provide more informative insights~\cite{howald-abramson-2012-use}, offering advantages for analysis and greater control over data.  In this context, the ability to provide granular topics is a highly desirable attribute of topic modeling for business use-cases. To confirm this as a preliminary study, in this work, we conduct an internal study with our business insight analysts where analysts choose their preferred topics amongst topics of different levels of granularity. Our results show that the business preferred medium to high levels of granularity 92\% of the time.\footnote{Details of the internal study are presented in Appendix~\ref{section:appendix:granularity_study}.} 


Despite its practical significance, the generation of granular topics has not received much attention. Most previous topic modeling studies have concentrated on discovering more coherent and diverse topics~\cite{bertopic, topicgpt, concept_induction, contextual_top2vec}, as well as on automatic topic naming~\cite{alokaili2020automatic}. Given the objectives of these studies, the topic modelling approaches have predominantly concentrated on higher-level concepts and have been assessed using classification datasets, which generally contain 10-20 high-level topics. Consequently, granular topic modeling remains an underexplored area that requires further investigation, given its importance in business applications.

In the light of this, we introduce a framework called TIDE (\textbf{T}opic \textbf{I}nsights and \textbf{D}ocument \textbf{E}xploration) that not only supports a novel \ac{LLM} based granular topic modeling approach but also offers various functions that facilitates better usability in real business use-cases, involving a targeted summarization, topic hierarchy detection, distillation, and automatic evaluation tool. Subsequently, we evaluate TIDE using three public datasets that are well-suited for granular topic modeling evaluation, as well as two real-world business datasets we encountered. Our experimental results demonstrate that TIDE is highly competent in granular topic modeling, surpassing modern topic modeling baseline approaches. While the significant variability is observed in traditional methods across various public datasets, TIDE consistently achieves high rankings regardless of dataset. Also, it offers various practical advantages for enhancing industrial use cases.

\begin{figure*}[ht]
    \centering
    \includegraphics[width=0.9\textwidth]{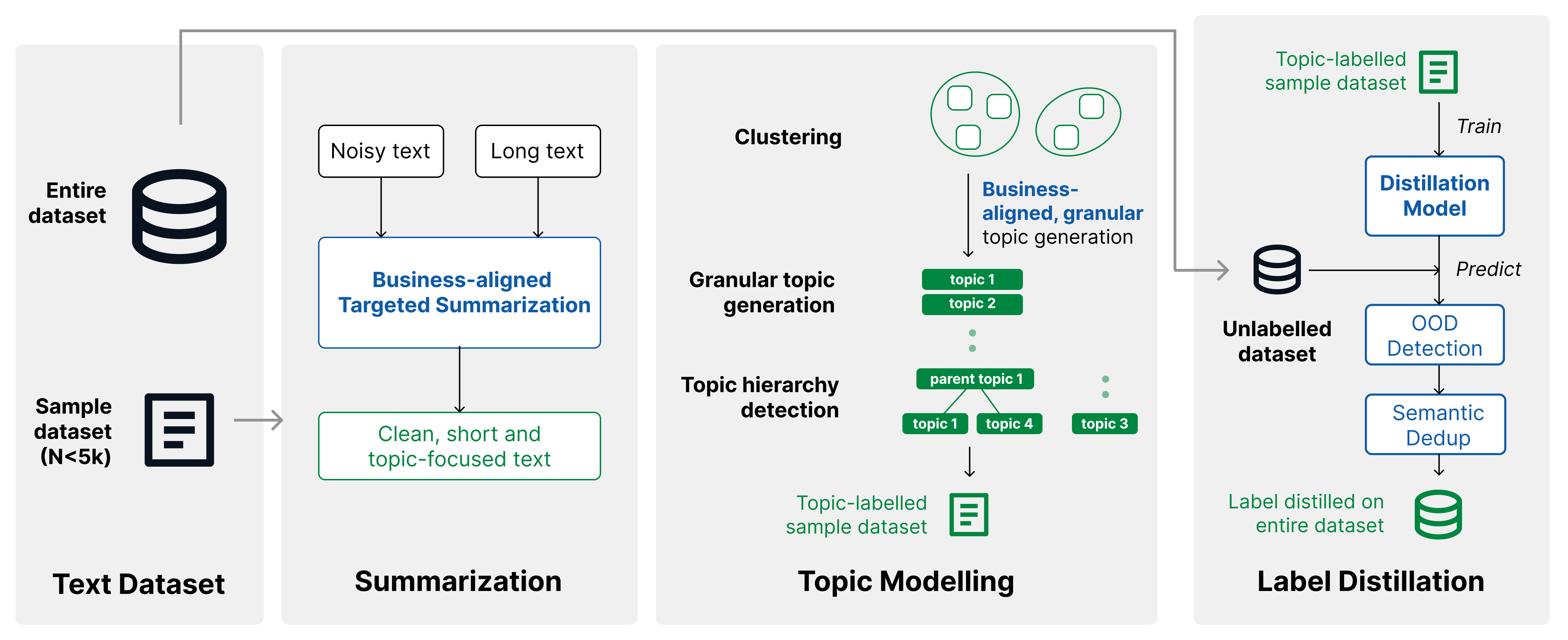}
    \vspace{-2ex}
    \caption{TIDE architecture diagram}
    \label{fig:tide_arch}
    \vspace{-2ex}
\end{figure*}

\section{TIDE Framework}
In this section, we present the TIDE framework, which has been developed as a Python library. Figure~\ref{fig:tide_arch} illustrates the overall TIDE architecture: with granular topic modelling at its core, along with supports for summarization for long documents and distillation for inference based on topic modelling output. More detailed figure with a specific example is presented in Figure~\ref{sec:appendix:tide_modules} in Appendix. LLMs are used in various library components and the prompts employed are detailed in Appendix~\ref{appendix:summarization_prompt} through \ref{appendix:hierarchy_detection_prompt}.

\subsection{Summarization}
\label{summarization_module}
We provide a targeted summarization module for datasets with lengthy, noisy, and complex-structured input texts, such as conversations logs or news articles, which are commonly encountered in real-business contexts. For such long and complicated texts, a pre-procesing step is required because the texts may exceed the \ac{LLM}s' context length, or more subtly, exceed the \ac{LLM}s' effective context length~\cite{hsieh2024ruler,liu-etal-2024-forgetting}. Additionally, preparing clean, concise, and well-structured text prior to topic modeling can alleviate \ac{LLM}s' performance issues associated with lengthy input contexts~\cite{lost_in_middle}. Although truncation can be considered, summarization is more desirable because it can retain key content relevant for topic generation.
Furthermore, our summarization module employs business' definition of desired topics to focus the summary for downstream topic modeling. For example, given customer-chatbot conversations, we can summarize both content that is explicitly mentioned (e.g. contact reason) and content that can be inferred implicitly (e.g. chatbot behavioral issues). This allows TIDE to derive business-relevant topics insights on both explicit and implicit content, enhancing usefulness to industry use-cases. The design of the summarization prompt is outlined in Appendix~\ref{appendix:summarization_prompt}.

\subsection{Granular Topic Modeling}
\label{sec:granular_topic_modelling}
TIDE topic modeling provides business-relevant, granular topics through a guided cluster-then-generate approach. First, TIDE clusters semantically similar texts using K-means clustering, which is a key for generating granular topics. Without clustering, \acp{LLM} will identify holistic features and are inclined to produce higher-level concepts. Conversely, clustering enables \acp{LLM} to access detailed information that might be considered insignificant in the perspective of the entire corpus and thus overlooked. Subsequently, each cluster of texts is provided to the \ac{LLM} to:(1) generate distinct, granular topic(s), and (2) accurately assign text to the appropriate topic(s). During topic generation, the clustering step is beneficial for several reasons. Firstly, it minimizes the \ac{LLM} costs. Secondly, by providing \ac{LLM} with similar texts, TIDE allows the \ac{LLM} to generate topics that generalize across the input texts so that the topics have the granularity tailored to the dataset. Additionally, TIDE can optionally incorporate business' definition of desired topics to maximize the usefulness for users (refer to Appendix~\ref{sec:appendix:business_defn} for examples of business' definitions).

When it comes to topic modeling, TIDE appears to follow procedures similar to LLooM~\cite{concept_induction}. However, TIDE differs from LLooM in several aspects, mainly in topic assignment process. First, LLooM scores each (\textit{document}, \textit{topic}) pair, and then allocates a topic with the highest score to the respective document. This process is resource-intensive and slow, as \acp{LLM} are invoked $O(N \times k)$ times, where $N$ and $k$ are the number of documents and the number of generated topics, respectively. Instead, TIDE assigns topics immediately after the clustering phase using a single \ac{LLM} call for each cluster. Consequently, TIDE reduces the number of \ac{LLM} calls to $O(N)$, enabling more resource-efficient topic assignment process. Second, the topic assignment process is optional in LLooM. However, we found that the topic assignment is a crucial feature in business use cases\footnote{Clients often request 1) access to the exact percentage of each topic 2) a manual examination of (\textit{document}, \textit{topic}) pairs.}, which led us to include the topic assignment process as a mandatory step. Third, before conducting topic modeling, LLooM preprocesses all documents via extractive summarization to capture their main gists, followed by abstractive summarization. However, we empirically found that allowing the \acp{LLM} to generate summaries or topics from content rich in details is crucial. Consequently, TIDE directly applies abstractive summarization to lengthy documents and recommends using raw inputs for shorter documents. Finally, similar to topicGPT~\cite{topicgpt}, TIDE incorporates a refinement module which tries to merge small clusters into prominent ones once all the topics are generated; LLooM does not conduct refinement.

\subsection{Topic Hierarchy Detection}\label{sec:topic_hierarchy_detection}
A user might find the generated topics to be excessively granular or may want to construct a hierarchical structure of the topics. TIDE provides further functionalities for identifying the topological relationships of granular topics. The cluster-then-generate step introduced in section \ref{sec:granular_topic_modelling} can be re-purposed to group similar granular topics into hierarchies. Taking the generated granular topics as inputs, the same algorithm performs cluster-then-generate using the prompt detailed in Appendix~\ref{appendix:hierarchy_detection_prompt}. After this process, for example, granular topics `Request for overdraft fee refund' and `Request for late fee refund' can be grouped under parent topic `Refund request'.


\subsection{Distillation}
Once topics are generated and assigned, most future documents can be classified into these existing topics, unless there is a change in the underlying topics~\cite{jang-etal-2024-driftwatch}. To facilitate cost effective classification, TIDE supports distillation by fine-tuning a pre-trained model (e.g. BERT) to learn to classify data into existing topics. Given that topics may originate from datasets with limited data, resulting in very few records per class, TIDE supports distillation using SetFit~\cite{tunstall2022efficientfewshotlearningprompts}, which has proven to be effective to learn classifications in situation of data scarcity. 

\subsection{Automatic Evaluation Tool}\label{sec:automatic evaluation}
Assessing the quality of generated topics following topic modeling is a natural step. However, most real-world data lacks labels, which hinders performance evaluation through comparison with ground-truth. In this regard, TIDE offers an automatic evaluation tool utilizing a \ac{LLM}, which measures the following three metrics.

\paragraph{Label accuracy. } Inspired by the work of \citet{mu2024addressing}, the label accuracy is designed to measure how distinct and appropriate the generated topic is compared to other topics. For each topic, we sample $N$ documents and identify the top-$k$ closest topics by using the embedding similarity. Next, we obtain the percentage where a \ac{LLM} retains the originally assigned topic when compared to the top-$k$ closest topics. Finally, the overall label accuracy is calculated as the macro average across all topics.

\paragraph{Topic accuracy/completeness. }  Motivated by recent research using \acp{LLM} as judges~\cite{zheng2023judging, verga2024replacing, gu2024survey}, we directly inquire a \ac{LLM} to assess the accuracy and completeness of the generated topic considering the original text on a scale ranging from 1 to 4. Our tool enables the incorporation of business definitions into evaluation prompts, aligning the judgment bias of \ac{LLM} with the perspective required by businesses. The prompt designs of these metrics are provided in Appendix~\ref{appendix:topic_accuracy} and \ref{appendix:topic_completeness}.

\begin{table}[t!]
    \begin{center}
    \renewcommand{\arraystretch}{1.0}
    \footnotesize{
        \centering{\setlength\tabcolsep{4.0pt}}
    }
    \begin{tabular}{cccc}
    \toprule
         \textbf{Dataset} & \textbf{Banking77} & \textbf{Bills} & \textbf{Wiki} \\ \midrule
        \# low-level topics  & 77 & 114 & 279 \\
         Avg. \# of words per text & 11 & 184 & 3K \\ \bottomrule
    \end{tabular}
    \caption{The basic statistics of the public datasets.}
    \label{table:pub_data_statistics}
    \end{center}
    \vspace{-4ex}
\end{table}

\begin{table*}[t!]
    \begin{center}
        \renewcommand{\arraystretch}{1.0}
        \footnotesize{
            \centering{\setlength\tabcolsep{2pt}}
        }
        \begin{tabular}{c|cccc|cccc|cccc}
        \toprule
        \multirow{2}{*}{\textbf{Models}} & \multicolumn{4}{c|}{\textbf{Banking77}} & \multicolumn{4}{c|}{\textbf{Bills}} & \multicolumn{4}{c}{\textbf{Wiki}} \\
        & P1 & ARI & NMI & $N_T$ & P1 & ARI & NMI & $N_T$ & P1 & ARI & NMI & $N_T$ \\ \hline
        BERTopic & \textbf{.705} & \underline{.468} & \textbf{.817} & 64 & .359 & .230 & .558 & 17 & .303 & .144 & .652 & 27 \\
        BERTopic w/ LLM Labeler & \underline{.686} & .444 & .769 & 61 & .343 & .195 & .537 & 16 & .292 & .120 & .633 & 26 \\
        C-Top2vec & .557 & .406 & .788 & 40 & .424 & .230 & \textbf{.767} & 279 & \textbf{.490} & \textbf{.341} & .685 & 90 \\
        C-Top2vec w/ LLM Labeler & .241 & .062 & .502 & 15 & .419 & .228 & \underline{.762} & 267 & .442 & .278 & .638 & 70 \\
        TopicGPT & .184 & .102 & .489 & 10 & .399 & .244 & .646 & 148 & .422 & .232 & .574 & 201 \\
        LLooM  & .111 & .037 & .322 & 6 & .159 & .048 & .295 & 6 & .110 & .046 & .424 & 9 \\ \hline
        TIDE w/o Summarization & .624 & \textbf{.486} & \underline{.800} & 83 & \textbf{.440} & \textbf{.285} & .711 & 292 & .389 & .259 & \textbf{.816} & 695 \\
        TIDE & - & - & - & - & \underline{.432} & \underline{.278} & .719 & 336 & \underline{.485} & \underline{.325} & \underline{.801} & 252 \\ 
        \bottomrule
        \end{tabular}
    \caption{The results from the experiments conducted on public datasets. $N_T$ denotes the number of generated topics. We report the average of five repetitions for each model and dataset. The best and second best performances are highlighted in bold and underlined, respectively.}
    \label{table:pub_data_experiments}
    \end{center}
    \vspace{-4.5ex}
\end{table*}

\section{Experiments on Public Datasets}

\subsection{Experiment Settings.}
\paragraph{Datasets.} To verify the competence of TIDE's granular topic modeling, we choose three public datasets containing low-level topics, which are well-suited for evaluating granular topic modeling: Banking77~\cite{Casanueva2020}, Bills~\cite{bills} and Wiki~\cite{DBLP:journals/corr/abs-1708-02182}. The basic statistics of the datasets are described in Table~\ref{table:pub_data_statistics}. We remarked that Bills and Wiki datasets are commonly used for topic modeling evaluations but on high-level topics~\cite{topicgpt, concept_induction}, and Banking77 dataset encompasses a specific industry domain: finance. Table~\ref{tab:granularity_dataset} in Appendix~\ref{sec:appendix:datasets} shows an example for each dataset with their high-level and low-level topics.

\paragraph{Evaluation metrics.} The public datasets contain ground-truth labels, enabling for the use of standard clustering-based evaluation metrics employed in previous studies~\cite{bills, topicgpt}. Therefore, we calculated Harmonic Mean of Purity (P1), Adjusted Rand Index (ARI), and Normalized Mutual Information (NMI) between the predicted and ground-truth topics. The detailed descriptions regarding the reported metrics are provided in Appendix~\ref{section:appendix:pub_eval_metrics}.

\paragraph{Baseline models.} We compare our proposed granular topic modeling approach with the following recently proposed topic modeling methods: 

\begin{itemize}[nosep, leftmargin=*]
    \item \textbf{BERTopic}~\cite{bertopic}: This approach employs document embeddings derived from a \ac{PLM}. The embeddings are then clustered, and for each cluster, topic representations are generated using a class-based TF-IDF.
    
    \item \textbf{C-Top2Vec}~\cite{contextual_top2vec}: This method segments a document using a sliding window and represents it within multi-vetor representation using \acp{PLM}. Topic vectors are defined as the centroid of clusters, and document-topic distributions are determined based on the topic to which the segments belong. Finally, each topic is labeled with words and phrases that are close to the topic vectors.
    
    \item \textbf{TopicGPT}~\cite{topicgpt}: This \ac{LLM}-based approach consists of two stages: topic generation that identifies underlying topics given corpus, and topic assignment, which allocates the previously generated topics to each document. We adopted the TopicGPT option that generates low-level topics, which may be more suitable for granular topic modeling.
    
    \item \textbf{LLoom}~\cite{concept_induction}. This \ac{LLM}-based framework is designed to identify high-level concepts. The method involves three steps: \textsc{Distill}-\textsc{Cluster}-\textsc{Synthesize} steps, which condenses data through summarization, performs clustering, and identify the underlying high-level concepts. The \textsc{Distill} and \textsc{Synthesize} steps involve the use of \acp{LLM}.

\end{itemize}

Unlike TopicGPT and Lloom, BERTopic and C-Top2vec define a topic as a set of related words, which has a downside of lacking readability and interpretability. Hence, we additionally evaluated a variation of these two approaches, where a \ac{LLM}-based automatic topic labeler is applied on top of the set of words.\footnote{The prompt design is presented in Appendix~\ref{appendix:topic_labeling_prompt}.} We employed \texttt{GPT-4o} and \texttt{all-mpnet-base-v2} (mpnet) as backbone \ac{LLM} and \ac{PLM} across our experiments respectively. More detailed information regarding the implementation and hyperparameter settings are provided in Appendix~\ref{section:appendix:imp_detail_hparams}. TIDE with summarization was not performed on the Banking77 dataset because the text input length is notably short.

\subsection{Experimental Results}
\paragraph{Overall results. } Table~\ref{table:pub_data_experiments} presents the experimental results. Overall, TIDE consistently outperforms most baselines, achieving the best or second-best performance across the majority of datasets. On the contrary, the baseline approaches show significant performance variability across different datasets. To evaluate the significance of performance differences, we conducted statistical T-tests with a significance level of 0.05. 

In the Banking77 dataset, BERTopic achieves the best performance, with a statistically significant gap in P1 and NMI compared to TIDE. However, it does not perform well on the Bills and Wiki datasets, where TIDE significantly outperforms it across all metrics. In contrast, C-Top2vec exhibits comparable performance with TIDE on Bills and Wiki datasets. Specifically, in the Bills dataset, its NMI is higher than that of TIDE, but it reports a lower ARI, with no significant difference observed in P1. In the Wiki dataset, TIDE achieves better performance in NMI but there is no significant difference in the other metrics. However, C-Top2vec reports significantly lower performance in P1 and ARI on the Banking77 dataset. Unlike the two baselines, TIDE demonstrates consistently strong performance across all the three datasets. Moreover, BERTopic and C-Top2Vec have a qualitative disadvantage in labeling each topic as a set of words, rendering them unintuitive and less readable. We incorporated an additional \ac{LLM} auto-labeler to compensate this, but it was found to degrade the performance. We conjecture that a leading cause, as shown by $N_T$ values, is that the auto-labeler merges some topics sharing similar words, thereby deteriorating the quality of the topics. However, TIDE does not suffer from this disadvantage.

Regarding \ac{LLM}-based baselines, our proposed approach consistently outperforms TopicGPT and LLooM by a large margin across all metrics. It is worth noting that, despite having a similar framework, LLooM performs significantly worse than TIDE while generating far fewer topics. This suggests that the distinctions outlined in Section~\ref{sec:granular_topic_modelling} are effective and advantageous for granular topic modeling.

\paragraph{Effect of summarization. } Through experiments, we confirm the benefit of TIDE summarization in addressing lengthy text inputs. In Bills dataset, there is no significant difference with and without summarization. However, in Wiki dataset, which is approximately 16 times longer, incorporating summarization leads to statistically significant improvements in P1 and ARI metrics. In this regard, TIDE is advantageous for processing from real-industry domains, such as legal sector, which typically contains lengthy and complex-structured text~\cite{chalkidis-etal-2022-lexglue}.

\paragraph{TIDE generates more topics. } We observe that TIDE not only attains strong performance in terms of evaluation metrics, but it also generates far more topics than the baselines, indicating that TIDE is highly proficient in granular topic modeling. Notably, in the Wiki and Banking77 datasets, the number of generated topics is almost close to that of ground-truth labels. Moreover, provided the generated topics are too granular, users can identify higher-level topics using the topic hierarchy detection function outlined in section~\ref{sec:topic_hierarchy_detection}. These features bestow TIDE a substantial practical advantage for analyzing real business data.

\section{Experiments on Real Business Datasets}

\subsection{Experiment Settings.}

\begin{table}[t!]
    \begin{center}
    \renewcommand{\arraystretch}{1.0}
    \footnotesize{
        \centering{\setlength\tabcolsep{5.0pt}}
    }
    \begin{tabular}{ccc}
    \toprule
         \textbf{Dataset} & \textbf{CCC} & \textbf{CI} \\ \midrule
        \# instances  & 603 & 3K \\
         Avg. \# of words per text & 120 & 10 \\ \bottomrule
    \end{tabular}
    \vspace{-1ex}
    \caption{The basic statistics of the industry datasets.}
    \label{table:industry_data_statistics}
    \end{center}
    \vspace{-4.5ex}
\end{table}

\paragraph{Datasets.} To validate TIDE's performance in a real-world setting, we report performance on two industry datasets. The \textbf{C}ustomer \textbf{C}hatbot \textbf{C}onversation (CCC) dataset contains long, noisy text, including conversation messages and chatbot logs obtained from a consumer banking chatbot application. The \textbf{C}ustomer \textbf{I}ssues (CI) dataset contains a short summary describing the issue that a customer raises during their daily banking usage. Table~\ref{table:industry_data_statistics} provides basic statistics for each dataset. The business definition of desired topics for these industry datasets is defined in Appendix~\ref{sec:appendix:business_defn}. 

\paragraph{Evaluation metrics.} As the industry datasets have no ground-truth labels, we used the automatic evaluation tool provided by TIDE, which is outlined in Section~\ref{sec:automatic evaluation}. Regarding topic completeness and topic accuracy, we additionally conducted a human evaluation, where annotators were asked to measure these two metrics using the identical scale, eg., 1 to 4.
The details of the human evaluation process are outlined in Appendix~\ref{appendix:human_eval}.

\begin{table*}[t!]
    \begin{center}
        \renewcommand{\arraystretch}{1.0}
        \footnotesize{
            \centering{\setlength\tabcolsep{1pt}}
        }
        \begin{tabular}{cc|ccccc|ccccc}
        \toprule
        \multicolumn{2}{c|}{\multirow{2}{*}{\textbf{Models}}} & \multicolumn{5}{c|}{\textbf{CCC}} & \multicolumn{5}{c}{\textbf{CI}} \\
        & & $\mathcal{L}_A$ & $\mathcal{T}_A$ & $\mathcal{T}_C$ & $\mathcal{H}_A$ & $\mathcal{H}_C$ & $\mathcal{L}_A$ & $\mathcal{T}_A$ & $\mathcal{T}_C$ & $\mathcal{H}_A$ & $\mathcal{H}_C$ \\ \hline
        
        \multirow{2}{*}{\textbf{Topic Modeling}} 
        & Baselines & 0.33 & 3.70 & 1.95 & 1.01 & 1.01 & 0.56 & 2.93 & 2.31 & 1.36 & 1.32 \\
        
        & TIDE-business & \textbf{0.60} & \textbf{3.89} & \textbf{3.62} & \textbf{2.74} & \textbf{2.77} & \textbf{0.60} & \textbf{3.57} & \textbf{3.32} & \textbf{2.46} & \textbf{2.41} \\ \hline
        
        \multirow{2}{*}{\textbf{Distillation}} & SetFit-mpnet & - & - & - & - & - & 0.57 & 3.33 & \textbf{3.32} & - & - \\        
        & mpnet & - & - & - & - & - & 0.58 & 3.23 & \textbf{3.32} & - & - \\
        
        \bottomrule
        \end{tabular}
    \vspace{-1ex}
    \caption{The industry dataset experimental results. $\mathcal{L}_A$ denotes label accuracy. $\mathcal{T}_A$ and $\mathcal{T}_C$ refer to topic accuracy and topic completeness, respectively, where as $\mathcal{H}_A$ and $\mathcal{H}_C$ indicate the corresponding metric as evaluated by human annotators. The best performances is highlighted in bold.}
    \label{table:industry_data_experiments}
    \end{center}
    \vspace{-4ex}
\end{table*}

\paragraph{Baseline models. } Based on the results from the public dataset experiments, it is inferred that among the baselines, C-Top2vec demonstrates strong performance with lengthy text inputs, whereas BERTopic performs better with short text inputs. Hence, we selected C-Top2vec and BERTopic as a baseline for CCC and CI dataset, respectively. For both approaches, we incorporated the \ac{LLM} auto labeler (since their raw outputs, consisting of set of words, may pose potential disadvantages in \ac{LLM}- and human-driven evaluations).

\subsection{Experimental Results}


\paragraph{Topic modeling results.} The experimental results are summarized in Table~\ref{table:industry_data_experiments}. It is notable that TIDE greatly outperforms the baseline model across all metrics and dataset. Regarding our automatic evaluation metrics, the performance gap is very significant, particularly with an 85\% improvement in topic completeness, implying that TIDE generates more distinct and information-complete topics. In the human evaluation, TIDE also substantially outperforms baseline approaches, improving by 127\% on average. However, the scale of metrics is smaller than that of automated metrics, indicating human annotators exhibit a stricter and more conservative judgment bias compared to our LLM judges. Our results suggest that, TIDE outperforms previous methods not only on public datasets but proves to be more practical from an industry perspective, generating topics that are more granular yet accurate, informative, and distinctive.

\paragraph{Distillation results.} Additionally, we tested our distillation functionality with a new batch of CI dataset consisting of 1.8K instances. Compared to original TIDE topic modeling, we observe that the distillation model achieves slightly worse but almost comparable results in terms of automatic evaluation. For example, the label accuracy of mpnet without SetFit is 0.58, comparable to TIDE topic modeling label accuracy 0.60. The topic accuracy and topic completeness follow similar trends. SetFit performs similarly with vanilla fine-tuning approach; we expect the SetFit to work well in data-scarce real world scenarios.

\section{Related Works}
Topic modeling is an \ac{NLP} technique to automatically discover hidden semantic patterns underlying in a text corpus. Traditional statistical methods, such as \ac{LDA}~\cite{blei2003latent} and \ac{NMF}~\cite{fevotte2011algorithms}, assume that topics are latent variables between words and documents, and attempt to uncover document-topic and topic-word distributions. A significant downside of these methods is that they rely on word-count based bag-of-words representations, which overlook the semantic meanings and relationships between words. In an attempt to overcome this drawback, topic modeling approaches that utilize dense vector representations and clustering algorithm have been proposed~\cite{sia2020tired, top2vec, bertopic, contextual_top2vec}. Moreover, the recent success of \acp{LLM} has enabled researchers to apply the models for topic modeling by directly asking \acp{LLM} to identify/refine topics underlying a text corpus~\cite{topicgpt, concept_induction}. A distinctive feature of \ac{LLM}-based methods is their capability to generate interpretable and human-readable topics, unlike the aforementioned approaches where topics are represented as a set of related related words. Unlike the previously mentioned studies that primarily focused on generating higher-quality topics, hierarchical topic modeling placed greater emphasis on identifying the manifold structure of topics~\cite{viegas2020cluhtm, isonuma2020tree, chen2021hierarchical, chen2023nonlinear, wu2024affinity}.

\section{Conclusion}
From a business perspective, providing granular topics is essential as they offer deeper insights and more detailed information, thereby enhancing their practical utility. Nevertheless, most of prior studies regarding topic modeling have concentrated on uncovering high-level latent concepts rather than granular topics. In this paper, we introduce TIDE, which encompasses a novel topic modeling approach for identifying granular concepts, along with various practical components that are useful for addressing business scenarios. Our experimental results indicate that TIDE outperforms recent baseline approaches in granular topic modeling, and its auxiliary functionalities offer great assistance for dealing with industrial scenarios.

\newpage
\section*{Limitations}
Our approach is limited by the context length limitations of \acp{LLM}, making it infeasible to process lengthy texts that exceed the maximum token capacity. Although we employ summarization to condense each data point, we are still only able to feed data that \ac{LLM}s' context length allows. This may lead to information loss and prevent us from leveraging the full power of the data. 

Additionally, our distillation module utilizes the outputs of topic modeling outputs as a gold-label. As a result, its performance may decline if the topic modeling results contain ambiguous data points or if too may topics are generated with imbalanced distributions. Addressing these limitations will be a focus for our future efforts for improvement.

\section*{Disclaimers}

This paper was prepared for informational purposes by the Machine Learning Center of Excellence group of JPMorgan Chase \& Co. and its affiliates (``JP Morgan''). JP Morgan makes no representation and warranty whatsoever and disclaims all liability, for the completeness, accuracy or reliability of the information contained herein. This document is not intended as investment research or investment advice, or a recommendation, offer or solicitation for the purchase or sale of any security, financial instrument, financial product or service, or to be used in any way for evaluating the merits of participating in any transaction, and shall not constitute a solicitation under any jurisdiction or to any person, if such solicitation under such jurisdiction or to such person would be unlawful.


\begin{acronym}
  \acro{NLP}{Natural Language Processing}
  \acro{LLM}{large language model}
  \acro{PLM}{pre-trained language model}
  \acro{LDA}{Latent Dirichlet Allocation}
  \acro{NMF}{Non-negative Matrix Factorization}
\end{acronym}
\bibliography{industry_paper_reference}

\clearpage
\appendix
\section{Appendix}

\subsection{Internal study on topic granularity}
\label{section:appendix:granularity_study}

To quantify business preference on topic granularity, we conducted an internal study with 2 business insights analysts working on a chatbot used in a Bank. Their primary goal for topic insight generation is to understand key failure points in the customer journey and identify high-impact fixes to improve customer experience. 

Specifically, the analysts were asked to review 50 samples of real customer feedback about the chatbot. The annotators were asked to choose the best topic representation out of 3 options that contained varying levels of granularity. Note that the topic options were shuffled in order to reduce annotator bias. Below are the instructions provided to the analysts for the annotation task.

\medskip
\textit{As an insights analyst for the a chatbot used in a Bank, your role is to analyze customer reviews and identify high-impact ways to improve customer experience.  You are interested in identifying the following:}

\begin{itemize}
    \item \textit{What are some major issues or pain-points that exist in the chatbot experience?}
    \item \textit{Which features or customer journeys are causing issues? How can they be fixed?}
    \item \textit{What features or journeys work well for users? What do they like about them?}
\end{itemize}

\textit{To answer these questions, a topic model can be used to identify topics present in customer reviews. Topics discovered by the topic model will reveal common customer feedback, complaints or issues that are mentioned in the customer reviews. By quantifying prevalent topics, you will be able to identify high-impact improvements to address major concerns and understand key insights about the chatbot.}
 
\textit{Given the above context, we would like to understand what ‘granularity of information’ you would prefer to preserve in a topic representation.  }
 
\textit{You are provided 50 customer reviews to annotate. Each customer review has tagged with topics of varying granularities (topic 1, topic 2,  topic 3). Your goal is to identify which topic granularity is most useful for insight generation for the given text.}
\medskip

Table \ref{tab:granularity_study_example} shows an example of a customer review, along with the topic options that were provided. The outcome of this study shows that analysts preferred highly granular topics 56\% of the time, medium granular topics 36\% of the time, and broad topics 5\% of the time. Ultimately, the study confirmed that topics with higher granularity are more informative and helpful for business insights. 

\begin{table}[t]
    \begin{center}
    \renewcommand{\arraystretch}{1.3}
    \footnotesize{
        \centering{\setlength\tabcolsep{2.0pt}}
    }
    \begin{tabular}{>{\centering\arraybackslash}m{1.8cm}>{\arraybackslash}m{5cm}}
        \toprule
        \centering\textbf{Customer Review} & I had nothing to do with suspicious fraud activity and you close my account on me and I'm stranded with no money in Tennessee \\ \midrule

        \centering\textbf{Highly Granular} & Account closure following incorrect detection of fraud activity \\ \midrule
        \centering\textbf{Medium Granular} & Fraud and account closure \\ \midrule
        \centering\textbf{Broad} & Fraud \\
        
        \bottomrule
    \end{tabular}
    \caption{Example of customer review with topic granularity options.}
    \label{tab:granularity_study_example}
    \end{center}
    \vspace{-4ex}
\end{table}


\subsection{Evaluation Metrics for Public Dataset Experiments}~\label{section:appendix:pub_eval_metrics}

For the public datasets, which contain ground-truth topics, we used clustering-based evaluation metrics in accordance with previous studies. Suppose we have $N$ instances and two clusterings $U=\{u_1, u_2, ..., u_N\}$ and $V=\{v_1,v_2,...,v_C\}$, where $u_i$ and $v_i$ represent the cluster assigned to $i$-th instance by $U$ and $V$, respectively. Here, $u_i \in\{U_1,U_2,...,U_K\}$ and $v_i \in \{V_1,V_2,...,V_C\}$.

\paragraph{Harmonic Mean of Purity.} Purity is an intuitive metric that measures the number of ground-truth labels present among in all documents that belong to a single predicted cluster; it is analogous to precision~\cite{zhao2001criterion}. Intuitively, a small number of ground-truth labels within a predicted cluster suggests a high alignment between the identified concepts and the ground-truth concepts~\cite{bills}. The purity is computed as follows,  with $V$ considered as the ground-truth:
\begin{align*}
    P(U,V)=\frac{1}{N} \sum_{k} \max_c |C_{U_k} \cap T_{V_c}|,
\end{align*}
where $C_{U_k}$ represents the set of instances assigned to cluster $U_k$, and $T_{V_c}$ denotes the set of instances in ground-truth label $V_c$. However, a limitation of purity is that it is not a symmetrical metric. Consequently, the Harmonic Mean of Purity (P1) is commonly used in practice, which is calculated as follows:
\begin{align*}
    P1(U,V)=\frac{2 \times P(U,V) \times P(V,U)}{P(U,V) + P(V,U)}.
\end{align*}

\paragraph{Adjusted Random Index.} ARI is a modified metric of Random Index (RI,~\citealt{RI}). The RI is a metric used to assess whether two clustering results are agree or disagree, which is given by:
\begin{align*}
    RI(U,V)= (N_{00} + N_{11})/{\binom{N}{2}},
\end{align*}
where $N_{00}$ and $N_{11}$ represent the number of pairs that are in different clusters and same clusters, respectively, in both U and V. However, the RI has some limitations, such as, in practice, its values often fall within the narrow range from 0.5 to 1 in practice, even though its theoretical range is from 0 to 1~\cite{10.1145/1553374.1553511}. In this regard, the ARI is more commonly used, which adjusts the RI for the change grouping of elements~\cite{steinley2004properties}. The ARI is calculated as follows:
\begin{align*}
    ARI(U,V)=\frac{RI - RI_{\mathrm{E}}}{\max(RI) - RI_{\mathrm{E}}},
\end{align*}
where $RI_{\mathrm{E}}$ is the expected value of RI for random clusterings and $\max(R1)$ denotes the maximum value of the RI. The ARI ranges from -1 to 1, where 1 indicates perfect agreement between the two clusterings, 0 implies random cluster assignments, and negative values indicate less agreement than would be expected by chance~\cite{ARI, 10.1145/1553374.1553511}.

\paragraph{Normalized Mutual Information.} NMI is a metric designed to assess the similarity between two clustering results based on the concept of mutual information (MI, \citealt{shannon1948mathematical}), which quantifies the amount of information shared between two random variables. As the name implies, the NMI normalizes the MI to a value between 0 and 1, reducing the sensitivity of the MI to variations in the number of clusters~\cite{strehl2002cluster}. The NMI is calculated as follows:
\begin{align*}
    NMI(U,V)=\frac{2 \times \mathcal{I}(U,V)}{\mathcal{H}_U + \mathcal{H}_V},
\end{align*}
where $\mathcal{I}(U,V)$ is the MI between the clusterings $U$ and $V$. $\mathcal{H}_U$ and $\mathcal{H}_V$ represent the entropies of the clusterings $U$ and $V$, respectively.

\subsection{Implementation Details of Baseline Models}~\label{section:appendix:imp_detail_hparams}
\paragraph{Contextual-Top2vec. } For implementation, we used the original code shared by the authors.\footnote{\href{https://github.com/ddangelov/Top2Vec}{https://github.com/ddangelov/Top2Vec}} In accordance with the paper, we employed \texttt{all-mpnet-base-v2} as the backbone embedding model and applied the same hyperparameter settings as the authors, which are detailed in Table 8 of their paper~\cite{contextual_top2vec}.

\paragraph{LLooM. } For implementation, we used the official code shared by the authors.\footnote{\href{https://github.com/michelle123lam/lloom}{https://github.com/michelle123lam/lloom}} In accordance with the paper, we employed \texttt{all-mpnet-base-v2} as the backbone embedding model for clustering. We used the full pipeline exposed by $gen\_auto$ API in all experiments with the default parameter settings. To choose a single topic to compare with other approaches, the topic with max score was chosen.

\paragraph{BERTopic. } For implementation, we used the public BERTopic library.\footnote{\href{https://github.com/MaartenGr/BERTopic}{https://github.com/MaartenGr/BERTopic}} In accordance with the paper, we employed \texttt{all-mpnet-base-v2} as the backbone embedding model for BERTopic embedding step. We used BERTopic $fit$ API with its default configurations to conduct our experiments.

\paragraph{TopicGPT} For implementation, we used the original code shared by the authors.\footnote{\href{https://github.com/chtmp223/topicGPT}{https://github.com/chtmp223/topicGPT}} We used the full pipeline including \texttt{refinement} to conduct our experiments. To make fair comparison with other methods, we generated and evaluated on low-level topic generations exposed by $generate\_topic\_lvl2$ API in TopicGPT. 

\subsection{Human Annotation Process. }\label{appendix:human_eval}
Topic accuracy and topic completeness are metrics that depend entirely on the judgment of \acp{LLM}, which may misaligned with human perspectives. Therefore, we also performed human evaluation for these two automatic metrics. For each dataset, we selected a sample of 200 input text instances along with their corresponding topics, which are generated by TIDE and the baseline approaches. Subsequently, the samples are given to our in-house four anglophone annotators for evaluation. The prompt outlined in Appendix~\ref{appendix:topic_accuracy} and \ref{appendix:topic_completeness}, which serve as a prompts for the automatic metrics, are also provided to annotators as evaluation guidelines. During the evaluation, we did not display the name of model used to generate the topic and shuffled the topics to prevent any potential bias in decision-making.

\section{Prompt Designs}
This section introduces the prompt designed used for the TIDE framework and experiments. The placeholders enclosed in curly brackets (\{\}) are where an actual text data is inserted.

\subsection{Summarization Prompt}~\label{appendix:summarization_prompt}
The following prompt is used for \acp{LLM} to generate a business-aligned targeted summary of a length document.

\vspace{5mm}
\noindent \texttt{You are a professional top-performing data analyst that generates insights and reports to support decision-making and improve business processes.}

\vspace{2mm} 
\noindent \texttt{You are reviewing TEXT in the following domain: \{domain\_description\}}

\vspace{2mm} 
\noindent \texttt{Use the following `Summary Guidelines` to generate a summary:}

\vspace{1mm} 
\noindent \texttt{- The SUMMARY must contain information that relates to the TOPIC DESCRIPTION and TOPIC DEFINITION.}

\vspace{1mm} 
\noindent \texttt{- The SUMMARY must contain specific keywords and important phrases that were mentioned in the TEXT, if it relates to the TOPIC DESCRIPTION and TOPIC DEFINITION.}

\vspace{1mm}
\noindent \texttt{- The SUMMARY may omit meaningless sentences in the TEXT, such as greetings.}

\vspace{1mm}
\noindent \texttt{- The SUMMARY MUST NOT contain any new information that is not present in the TEXT. DO NOT add any extra details or other texts.}

\vspace{1mm}
\noindent \texttt{- The SUMMARY is a re-writing of the content written in the TEXT.}

\vspace{1mm}
\noindent \texttt{- DO NOT answer any questions or try to respond to anything in the TEXT.}

\vspace{1mm}
\noindent \texttt{- Never generate URLs or links apart from those present in the TEXT.}

\vspace{1mm}
\noindent \texttt{- You CANNOT respond with videos, memes, photos, code, or other non-English language content.}

\vspace{1mm}
\noindent \texttt{- Your response should never contain toxic, or NSFW material. }

\vspace{2mm}
\noindent \texttt{TOPIC DESCRIPTION: \{topic\_description\}}

\vspace{2mm}
\noindent \texttt{TOPIC DEFINITION: \{topic\_definition\}}

\vspace{2mm}
\noindent \texttt{TEXT: \{text\}}

\vspace{2mm}
\noindent \texttt{\{format\_instructions\}}

\vspace{2mm}
\noindent \texttt{SUMMARY:}

\subsection{Topic Generation Prompt}~\label{appendix:topic_generation_prompt}
To generate topics from given documents, we first use the following prompt.

\vspace{5mm}
\noindent \texttt{Your task is to generate topics within the documents.}

\vspace{2mm}
\noindent \texttt{\{topic\_description\}}

\vspace{2mm}
\noindent \texttt{\{topic\_generation\_examples\}}

\vspace{2mm}
\noindent \texttt{[Instruction]}

\vspace{1mm}
\noindent \texttt{Determine topics mentioned in the document.}

\vspace{1mm}
\noindent \texttt{- The topics must reflect a SINGLE topic instead of a combination of topics.}

\vspace{1mm}
\noindent \texttt{- \{topic\_definition\}}

\vspace{1mm}
\noindent \texttt{- \{no\_topic\_option\}}

\vspace{1mm}
\noindent \texttt{- If there are several topics, simple separate them with comma: <topic1>, <topic2>, ...}

\vspace{1mm}
\noindent \texttt{Document: \{document\}}
\vspace{5mm}

Subsequently, we refine the generated topics by merging smaller clusters into a more prominent cluster using the prompt provided below.

\vspace{5mm}
\noindent \texttt{You will receive a list of topics that belong to the same level of a topic hierarchy. Your task is to merge topics that are paraphrases or near duplicates of one another. Return "None" if no modification is needed. 
}

\vspace{2mm}
\noindent \texttt{Here are some examples:}

\vspace{2mm}
\noindent \texttt{[Example 1: Merging topics]}

\vspace{1mm}
\noindent \texttt{\{topic\_merge\_examples\}}

\vspace{2mm}
\noindent \texttt{[Rules]}

\vspace{1mm}
\noindent \texttt{- Each line represents a topic that is descriptive. }

\vspace{1mm}
\noindent \texttt{- \{topic\_definition\}}

\vspace{1mm}
\noindent \texttt{- Perform the following operations as many times as needed: }

\vspace{1mm}
\setlength{\leftskip}{10pt} \noindent \texttt{- Merge relevant topics into a single topic.}

\vspace{1mm}
\setlength{\leftskip}{10pt} \noindent \texttt{- Do nothing and return `None' if no modification is needed.}              

\vspace{1mm}
\noindent \texttt{- Return each merged topic in its own line, along with the original indices of the topics that were merged to this topic separated by a comma.}

\vspace{2mm}
\noindent \texttt{[Topic List]}

\vspace{1mm}
\noindent \texttt{\{topic\}}

\vspace{2mm}
\noindent \texttt{Output the modification or `None' where appropriate. Do not output anything else.}

\vspace{1mm}
\noindent \texttt{[Your response]}

\subsection{Topic Assignment Prompt}~\label{appendix:assignemnt_prompt}
The following prompt is designed for \acp{LLM} to categorize a document under a specific topic from a pre-defined list.

\vspace{5mm}
\noindent \texttt{You will receive a document that very probably fall into one of the main topics. Return the main topic that the topic belongs to. If no main topic is appropriate, return `Other'.} 

\vspace{2mm}                      
\noindent \texttt{[Document]}
                        
\noindent \texttt{\{document\}}

\vspace{2mm}                      
\noindent \texttt{[Main topics]}
                        
\noindent \texttt{\{main\_topics\}}

\vspace{2mm}                      
\noindent \texttt{[Your response]}

\subsection{Topic Hierarchy Detection Prompt}~\label{appendix:hierarchy_detection_prompt}
The prompt below is used for identify hierarchical relationships existing in the topological connections of the generated topics.

\vspace{5mm}
\noindent \texttt{You will receive a list of topics that belong to the same level of a topic hierarchy. Your task is to group topics into broad parent topics. } 

\vspace{2mm}
\noindent \texttt{Here are some examples: }

\vspace{2mm}
\noindent \texttt{[Example 1: Generate parent topics]}

\vspace{2mm}
\noindent \texttt{\{parent\_topic\_examples\}}

\vspace{2mm}
\noindent \texttt{[Rules]}

\vspace{1mm}
\noindent \texttt{- Each line represents a topic that is descriptive. }

\vspace{1mm}
\noindent \texttt{- Perform the following operations as many times as needed: }

\vspace{1mm}
\setlength{\leftskip}{10pt} \noindent \texttt{- Create a general / broad / parent topic for the topics groups identified.}

\vspace{1mm}
\noindent \texttt{- Return each parent topic in its own line, along with the original indices of the topics that belong to this topic separated by a comma.}

\vspace{2mm}
\noindent \texttt{[Topic List]}

\vspace{1mm}
\noindent \texttt{\{topics\}}

\vspace{2mm}
\noindent \texttt{[Your response]}

\subsection{Automatic Topic Labeling Prompt}~\label{appendix:topic_labeling_prompt}
We employed the following prompt for automatically naming topics based on the set of words generated by the BERTopic and C-Top2vec approaches.

\vspace{5mm}
\noindent \texttt{Generate a human-readable topic that encompasses a given set of words.
Consider the relationships, common themes, or overarching concepts that link the words together. Ensure the topic is broad enough to cover all the words but specific enough to be meaningful.}

\vspace{2mm}                      
\noindent \texttt{\# Output Format}

\noindent \texttt{Provide the topic as a short, concise phrase.}

\vspace{2mm}                      
\noindent \texttt{Set of Words: \{set\_of\_words\}}

\vspace{2mm}
\noindent \texttt{[Your response]}

\subsection{AutoEval: Topic Accuracy Prompt}~\label{appendix:topic_accuracy}

\paragraph{CCC dataset prompt. } The following prompt is used to measure topic accuracy of topics generated from the CCC dataset.

\vspace{3mm}
\noindent \texttt{Your task is to evaluate the accuracy of the topic over the given text; how accurate the topic is and well represented in the given text. You can ignore the level of granularity / detail. Please just assess if the topic is accurate given the text in any level of detail.}

\vspace{2mm}                      
\noindent \texttt{You should evaluate the topic accuracy in four level: Incorrect, Partially Correct, Mostly Correct, and Completely Correct.}

\vspace{2mm}                      
\noindent \texttt{Refer to the following business definitions when evaluating:}

\vspace{1mm}                      
\noindent \texttt{- topic definition: Each topic should be descriptive and help us understand different contact reasons.}

\vspace{1mm}                      
\noindent \texttt{Topics MUST contain the following details if exists:}

\vspace{1mm}                      
\noindent \texttt{- Specific names of [BANK\_NAME] accounts or services related to the customers reason for contacting.}
\noindent \texttt{- Specific problems or issues that the customers had.}
\noindent \texttt{- Specific feedback that customers have about [BANK\_NAME].}

\paragraph{CI dataset prompt. } To measure the topic accuracy of topics generated from CI dataset, the prompt is similar to that used for the CCC dataset, but it incorporates different business definitions, as outlined below:

\vspace{2mm}                      
\noindent \texttt{Refer to the following business definitions when evaluating:}

\vspace{1mm}                      
\noindent \texttt{- topic definition: Each topic should be very concise on the type of problems that the customers face.}

\vspace{1mm}                      
\noindent \texttt{- domain description: You are reading summary of various issues a bank customer face.}

\subsection{AutoEval: Topic Completeness Prompt}~\label{appendix:topic_completeness}

\paragraph{CCC dataset prompt. } The following prompt is used to measure topic completeness of topics generated from the CCC dataset.

\vspace{3mm}
\noindent \texttt{Your task is to evaluate the completeness of the topic, indicating the extent to which all necessary and relevant information is included. The complete topic means:}

\vspace{2mm}                      
\noindent \texttt{1) The topic covers all the necessary information presented in the text, suggesting that there are no missing topics.}

\vspace{2mm}                      
\noindent \texttt{2) The topic delivers an optimal level of information, meaning that the scope it covers does not contain unrelated information in relation to the text.}

\vspace{2mm}                      
\noindent \texttt{You should evaluate the topic completeness in four level: Not covered, Minorly covered, Mostly covered, and Complete.}

\vspace{2mm}                      
\noindent \texttt{Refer to the following business definitions when evaluating:}

\vspace{1mm}                      
\noindent \texttt{- topic definition: Each topic should be descriptive and help us understand different contact reasons.}

\vspace{1mm}                      
\noindent \texttt{Topics MUST contain the following details if exists:}

\vspace{1mm}                      
\noindent \texttt{- Specific names of [BANK\_NAME] accounts or services related to the customers reason for contacting.}
\noindent \texttt{- Specific problems or issues that the customers had.}
\noindent \texttt{- Specific feedback that customers have about [BANK\_NAME].}

\paragraph{CI dataset prompt. } To measure the topic completeness of topics generated from CI dataset, the prompt is similar to that used for the CCC dataset, but it incorporates different business definitions, as outlined below:

\vspace{2mm}                      
\noindent \texttt{Refer to the following business definitions when evaluating:}

\vspace{1mm}                      
\noindent \texttt{- topic definition: Each topic should be very concise on the type of problems that the customers face.}

\vspace{1mm}                      
\noindent \texttt{- domain description: You are reading summary of various issues a bank customer face.}

\clearpage
\onecolumn
\section{Examples}~\label{sec:appendix:datasets}

\begin{table*}[h]
\centering
\begin{tabular}{m{6.5cm}m{3cm}m{3cm}c}
    \toprule
    \makecell[c]{\textbf{Text}} & \makecell[c]{\textbf{High-level topic}} & \makecell[c]{\textbf{Low-level topic}} & \makecell[c]{\textbf{Dataset}} \\
    \midrule
    How do I add my new card? & \makecell[c]{card} & \makecell[c]{card linking} & banking77 \\ 
    \midrule
    
    Amends the Harmonized Tariff 
    Schedule of the United States to extend ... & \makecell[c]{Foreign Trade} & \makecell[c]{Tariff and Imports} & Bills \\
    \midrule

    2005 USC Trojans football team = The 2005 USC Trojans football team represented ... & \makecell[c]{Sports and \\ recreation} & \makecell[c]{American college \\football} & Wiki \\
    \bottomrule
\end{tabular}
\caption{Granular topic modeling examples from each dataset. While previous topic modeling evaluations use high level topics, our task evaluates on more granular low level topics.}
\label{tab:granularity_dataset}
\end{table*}

\section{Business definition of desired topics for industry datasets}~\label{sec:appendix:business_defn}
\begin{table}[h]
    \begin{center}
    \renewcommand{\arraystretch}{1.3}
    \footnotesize{
        \centering{\setlength\tabcolsep{2.0pt}}
    }

    \begin{tabular}{>{\centering\arraybackslash}m{3.0cm}>{\arraybackslash}m{3.5cm}>{\arraybackslash}m{8.0cm}}
        \toprule
        \textbf{Dataset} & \makecell[c]{\textbf{Domain description}} & \makecell[c]{\textbf{Topic definitions}} \\ \hline
        Customer Chatbot Conversation & You are reading description of various reasons why Bank customers used the chatbot to reach out to Bank. & Each topic should be descriptive and help us understand different contact reasons. \newline Topics MUST contain the following details, if it exists: \newline
    - Specific names of Bank accounts or services related to the customers reason for contacting. \newline
    - Specific problems or issues that the customers had. \newline
    - Specific feedback that customers have about Bank.\\ \midrule
    
    Customer Issues & You are reading summary of various issues a bank customer face. & Each topic should be very concise on the type of problems that the customers face. \\ \bottomrule
    \end{tabular}
    \end{center}
    \caption{Business definition of desired topics}
    \label{tab:example}
\end{table}

\section{In-depth illustrative diagram of TIDE modules}~\label{sec:appendix:tide_modules}

\begin{figure*}[ht]
    \centering
    \includegraphics[width=0.95\textwidth]{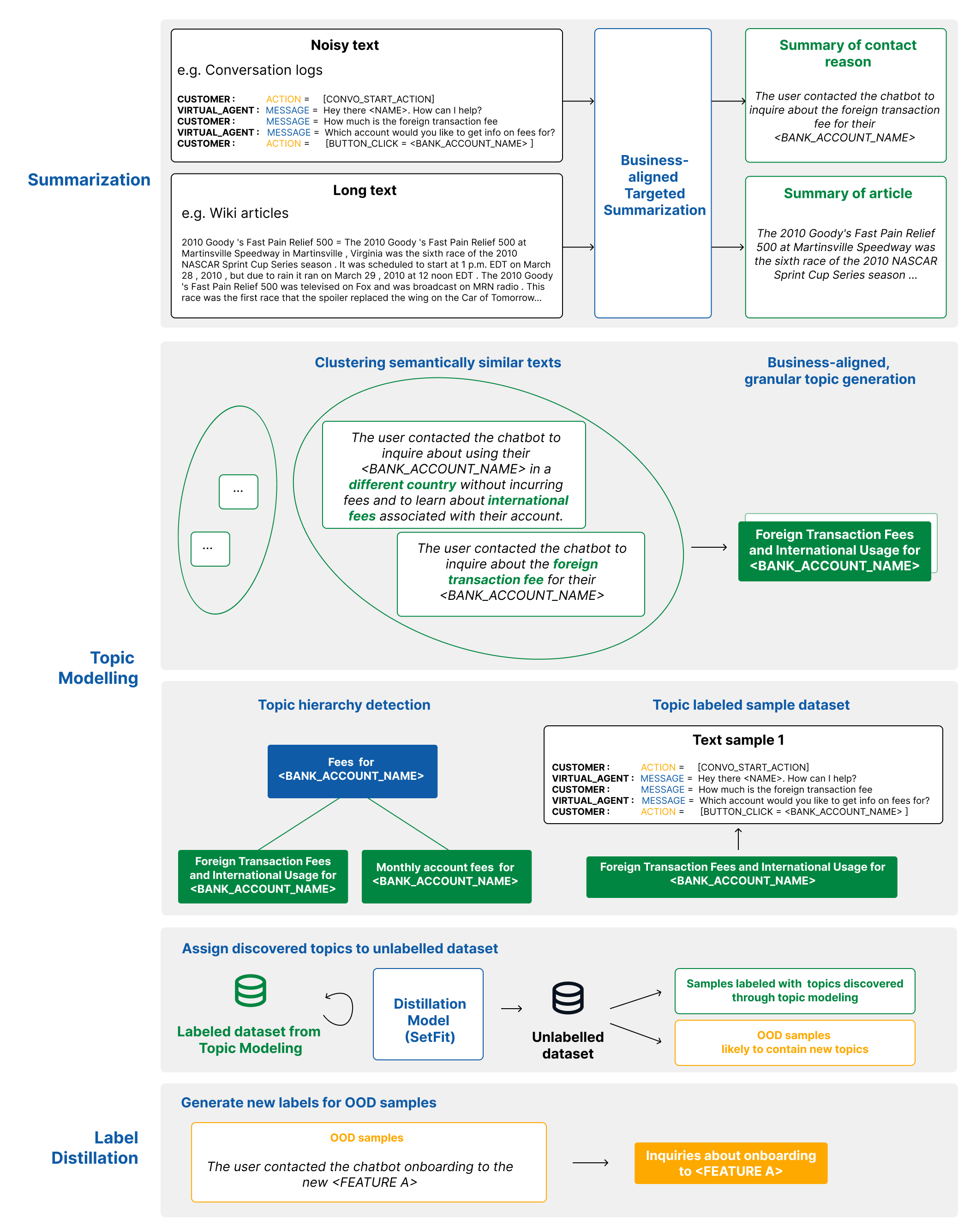}
    \vspace{-2ex}
    \caption{TIDE in-depth architecture diagram}
    \label{fig:tide_arch}
    \vspace{-2ex}
\end{figure*}

\end{document}